\documentclass[11pt, letterpaper, copyright, logo]{googledeepmind}

\usepackage{multirow}
\usepackage[authoryear, sort&compress, round]{natbib}
\usepackage{lipsum}
\usepackage{amsmath}
\usepackage{pstricks, pst-node}
\usepackage{verbatim}
\usepackage{multirow}
\usepackage{multicol}
\usepackage{longtable}
\usepackage{array}
\usepackage{listings, listings-rust}
\usepackage{booktabs}
\usepackage{enumitem}
\usepackage{xspace}
\usepackage{bm}
\usepackage{bbm}
\usepackage{mathtools}
\usepackage{soul}
\usepackage{graphicx}
\usepackage[most, breakable, skins]{tcolorbox}
\tcbuselibrary{skins}
\usepackage{subcaption}
\usepackage{amssymb}
\usepackage{colortbl}
\usepackage{csquotes}
\usepackage{setspace}
\usepackage[inkscapeformat=png]{svg}
\usepackage{tabularx,ragged2e}
\usepackage{makecell}
\usepackage{placeins}
\usepackage[symbol]{footmisc}
\usepackage[dvipsnames]{xcolor}
\usepackage{hyperref}
\usepackage{pifont}
\usepackage{cleveref}

\newcolumntype{L}[1]{>{\raggedright\let\newline\\\arraybackslash\hspace{0pt}}m{#1}}
\newcolumntype{C}[1]{>{\centering}m{#1}}

\newcolumntype{R}[1]{>{\raggedleft\let\newline\\\arraybackslash\hspace{0pt}}m{#1}}

\definecolor{ao}{rgb}{0.0, 0.0, 1.0}

\newcommand{\todo}[1]{}
\newcommand{\todor}[1]{}

\newcommand{\optimam}[1]{}

\newcommand\vcent[1]{\vcenter{\hbox{#1}}}
\newcommand\loudspeaker[1][3]{\ensuremath{\vcent{\rule{.6ex}{.6ex}}\kern-.5ex%
  \vcent{\scalebox{.6}[1]{\rotatebox[origin=center]{90}{$\blacktriangle$}}}%
  \ifnum#1>0\relax\kern.05ex\vcent{\scalebox{.4}{\ttfamily)}}%
  \ifnum#1>1\relax\kern-.4ex\vcent{\scalebox{.56}{\ttfamily)}}%
  \ifnum#1>2\relax\kern-.55ex\vcent{\scalebox{.7}{\ttfamily)}}%
  \fi\fi\fi}%
}

\title{Health AI Developer Foundations}

\author[$\ast$, 1]{Atilla P. Kiraly} 
\author[$\ast$, 1]{Sebastien Baur}
\author[1]{Kenneth Philbrick}
\author[1]{Fereshteh Mahvar}
\author[1]{Liron Yatziv}
\author[1]{Tiffany Chen}
\author[1]{Bram Sterling}
\author[1]{Nick George}
\author[1]{Fayaz Jamil}
\author[1]{Jing Tang}
\author[1]{Kai Bailey}
\author[1]{Faruk Ahmed}
\author[1]{Akshay Goel}
\author[1]{Abbi Ward}
\author[1]{Lin Yang}
\author[1]{Andrew Sellergren}
\author[1]{Yossi Matias}
\author[1]{Avinatan Hassidim}
\author[1]{Shravya Shetty}
\author[1]{Daniel Golden}
\author[2]{Shekoofeh Azizi}
\author[1]{David F. Steiner}
\author[1]{Yun Liu}
\author[1]{Tim Thelin}
\author[1]{Rory Pilgrim}
\author[1]{Can Kirmizibayrak}

\affil[1]{Google Research}
\affil[2]{Google DeepMind}
\affil[$\ast$]{Equal contributions.}
\affil[ ]{Correspondence: health-ai-foundations@google.com}

\begin{abstract}
Robust medical Machine Learning (ML) models have the potential to revolutionize healthcare by accelerating clinical research, improving workflows and outcomes, and producing novel insights or capabilities. Developing such ML models from scratch is cost prohibitive and requires substantial compute, data, and time (e.g., expert labeling). To address these challenges, we introduce Health AI Developer Foundations (HAI-DEF), a suite of pre-trained, domain-specific foundation models, tools, and recipes to accelerate building ML for health applications. The models cover various modalities and domains, including radiology (X-rays and computed tomography), histopathology, dermatological imaging, and audio. These models provide domain specific embeddings that facilitate AI development with less labeled data, shorter training times, and reduced computational costs compared to traditional approaches. In addition, we utilize a common interface and style across these models, and prioritize usability to enable developers to integrate HAI-DEF efficiently. We present model evaluations across various tasks and conclude with a discussion of their application and evaluation, covering the importance of ensuring efficacy, fairness, and equity. Finally, while HAI-DEF and specifically the foundation models lower the barrier to entry for ML in healthcare, we emphasize the importance of validation with problem- and population-specific data for each desired usage setting. This technical report will be updated over time as more modalities and features are added.  
\end{abstract}

\begin{document}
\maketitle

\section{Introduction}
\label{sec:intro}
Machine learning (ML) models, trained on diverse data ranging from genomic sequences to clinical images, have the potential to transform healthcare in applications ranging from accelerating drug discovery to enabling personalized diagnoses. In day-to-day clinical workflows, ML models have also been developed to help automate manual processes, assist with triage, diagnosis, or prognosis, and more, with the goal of helping improve quality of care or efficiency.

However, building robust ML models for these domains entails challenges. Development often requires large, labeled datasets, which are expensive and time-consuming to create and curate. Beyond cost, sharing these datasets across institutions is often restricted due to privacy and other considerations. Data scarcity, especially for rare conditions and underrepresented populations, further hinders dataset curation and limits generalizability. Finally, significant compute resources are often necessary to train large models or when utilizing large datasets, and clinical and technical modality-specific expertise, especially in the use of DICOMs for pathology and radiology, is often needed to correctly prepare data for ML models.

To help address these challenges, we present Health AI Developer Foundations (HAI-DEF), with the goal of catalyzing the development and adoption of AI in healthcare. HAI-DEF includes foundation models, tooling, recipes, and ready-to-use research endpoints. These resources were created to enable researchers and developers to both iterate on research ideas quickly and have a lower-friction path to incorporating AI in real-world use settings. In the initial phase, we offer research endpoints and open-weight models, enabling generation of high-quality embeddings from medical images (chest X-rays (CXR), histopathology patches, skin images, computed tomography (CT) images) and audio recordings (health acoustics like coughs and breaths). These embeddings offer a compact yet information rich representation of the given data. By leveraging these embeddings and tooling, researchers and developers can build robust models that perform well across diverse clinical settings with significantly less labeled data, shorten model training times, and reduce computational costs.

Table~\ref{tab:summary} provides an overview of the modalities and features offered per modality. Documentation and additional information is available at the \href{http://developers.google.com/health-ai-developer-foundations}{HAI-DEF developer site.}
  
\definecolor{lightblue}{RGB}{220,230,255}
\definecolor{lightpurple}{RGB}{230,220,255}
\definecolor{ct}{RGB}{201,218,248}
\definecolor{histo}{RGB}{217,210,233}
\definecolor{xrayblue}{RGB}{207,226,243}
\definecolor{lightorange}{RGB}{249,203,156}
\definecolor{audioc}{RGB}{207,226,243}

\begin{table}[h]
\centering
\caption{{\bf HAI-DEF: A Multimodal Platform for accelerating health AI.} HAI-DEF supports research across audio signals, radiology, histopathology, and dermatology, with more modalities planned.  Foundation models for each modality can be accessed via research endpoints or deployed using open weight models in open-source containers.  Per-modality tools, including a custom histopathology library and a code-free X-ray embedding interface, simplify integration and accelerate research.}
\begin{tabular}{>{\columncolor{white}}c >{\columncolor{white}}c >{\columncolor{white}}c >{\columncolor{white}}c}
\toprule
 & \textbf{Research} & \textbf{Open Model} & \textbf{Open Source}  \\
\textbf{Modality} & \textbf{Endpoint} & \textbf{Weights} & \textbf{Container}  \\
\midrule
\cellcolor{audioc} Audio & \cellcolor{audioc} \checkmark & \cellcolor{white}  & \cellcolor{white} \\
\rowcolor{lightblue}
\addlinespace[0.5cm]
\cellcolor{ct} Computed Tomography & \cellcolor{ct} \checkmark & \cellcolor{white} & \cellcolor{white}  \\
\rowcolor{lightpurple}
\addlinespace[0.5cm]
\cellcolor{histo} Histopathology & \cellcolor{histo} \checkmark & \cellcolor{histo} \checkmark & \cellcolor{histo} \checkmark  \\
\rowcolor{lightblue}
\addlinespace[0.5cm]
\cellcolor{lightorange} Dermatology & \cellcolor{lightorange} \checkmark & \cellcolor{lightorange} \checkmark & \cellcolor{lightorange} \checkmark  \\
\rowcolor{lightpurple}
\addlinespace[0.5cm]
\cellcolor{xrayblue}  X-ray (CXR) & \cellcolor{xrayblue} \checkmark & \cellcolor{xrayblue} \checkmark & \cellcolor{xrayblue} \checkmark \\
\bottomrule
\end{tabular}
\label{tab:summary}
\end{table}

\section{Models}
\label{sec:Models}

HAI-DEF encompasses multiple distinct models, each tailored to a specific use case and trained using advanced techniques on large, diverse datasets. The following list is a snapshot of our current progress, and our goal is to expand this model set in the near future.

HAI-DEF encompasses multiple distinct models, each tailored to a specific data modality and trained using various techniques on large, diverse datasets. The following list snapshots our current available models, which we expect to expand in the near future. More information about how to access these models can be found at the \href{http://developers.google.com/health-ai-developer-foundations}{HAI-DEF developer site} (for CXR, Path and Derm Foundation) and respective GitHub repositories for the research endpoints (\href{https://github.com/Google-Health/google-health/blob/master/health_acoustic_representations/README.md}{HeAR} and \href{https://github.com/Google-Health/imaging-research/tree/master/ct-foundation}{CT Foundation}).

\subsection{CXR Foundation}
CXR Foundation \citep{Sellergren2022} is a set of 3 models, all using an EfficientNet-L2 (\cite{Qizhe2020}) image encoder backbone. The three models learned representations of CXRs by leveraging both the image data and the clinically relevant information available in corresponding radiology reports.
\begin{enumerate}
  \item The original CXR Foundation is trained using supervised contrastive learning (SupCon,~\citet{Khosla2020}). This model is used for legacy support and comparison purposes.
  \item ELIXR-C~\citep{Xu2023ELIXR} is an image/text encoder trained with the CLIP method~\citep{Radford2021}. This model tends to perform better at zero-shot tasks, i.e. those that do not require further data for training.
  \item ELIXR-B~\citep{Xu2023ELIXR} is trained with the BLIP-2 method~\citep{Li2022} on a dataset of over 1,000,000 chest radiographs from 5 hospitals in India, and 4 hospitals in the USA. This model tends to perform better on downstream classification tasks.
\end{enumerate}

\subsection{Path Foundation}
Path Foundation~\citep{Lai2023} is a Vision Transformer (ViT)~\citep{Dosovitskiy2020} encoder for histopathology image patches trained with self-supervised learning (Masked Siamese Networks,~\citet{assran2022maskedsiamesenetworkslabelefficient}). It incorporates pathology-specific optimizations, including approaches to help learn stain-agnostic features and to generalize across patch magnifications. Training for this model utilized hematoxylin and eosin (H\&E) stained whole slide images (WSIs) from The Cancer Genome Atlas (TCGA)~\citep{GDCDataPortal}.

\subsection{Derm Foundation}
Derm Foundation~\citep{Rikhye2024} is a BiT ResNet-101x3~\citep{Kolesnikov2019} image encoder trained using a two-stage approach on over 16K natural and dermatology images~\citep{Liu2020}. First, contrastive learning (ConVIRT,~\citet{Zhang2020}) was applied on a large number of image-text pairs from the internet (\citet{Sun2017}). Second, the model was fine-tuned to identify dermatology conditions on a mix of datasets, including tele-dermatology and skin cancer datasets.

\subsection{HeAR}
HeAR~\citep{Baur2024} is a ViT audio encoder trained using a Masked Autoencoder (MAE) approach~\citep{He2021} on a large dataset of 313 million unlabelled, non-medical audio clips. The model learns to reconstruct masked spectrogram patches, capturing rich acoustic representations of health-related sounds like coughs and breathing patterns.

\subsection{CT Foundation}
CT Foundation provides embeddings suitable for downstream classification tasks. The underlying model is VideoCoCa~\citep{yan2023}, a video-text model designed for efficient transfer learning from 2D Contrastive Captioners (CoCa)~\citep{yu2022}. CT Foundation was trained on CT volumes and radiology reports from over 500,000 cases across different anatomic regions, using a similar approach to \citet{yang2024}.

\section{Evaluations}
\label{sec:results}
\begin{figure}[htpb]
  \caption{Data efficiency comparison of four of our foundation models against established approaches. CXR (upper left) shows the average performance across six binary classification tasks with the original CXR Foundation model and the new ELIXR-B model; for Pathology (upper right), it focuses on a 4-class prostate cancer Gleason grading task for needle core biopsies and the detection of metastatic breast cancer in lymph nodes. For Derm (lower left), the evaluation centered on skin condition category (28-way) identification, while for HeAR (lower right), it involved identifying COVID-19 from cough sounds. Notably, the foundation models achieve comparable results while using substantially less training data and compute. CT Foundation is covered below; additional methods not available for equivalent comparisons.}
  {\includegraphics[width=\linewidth]{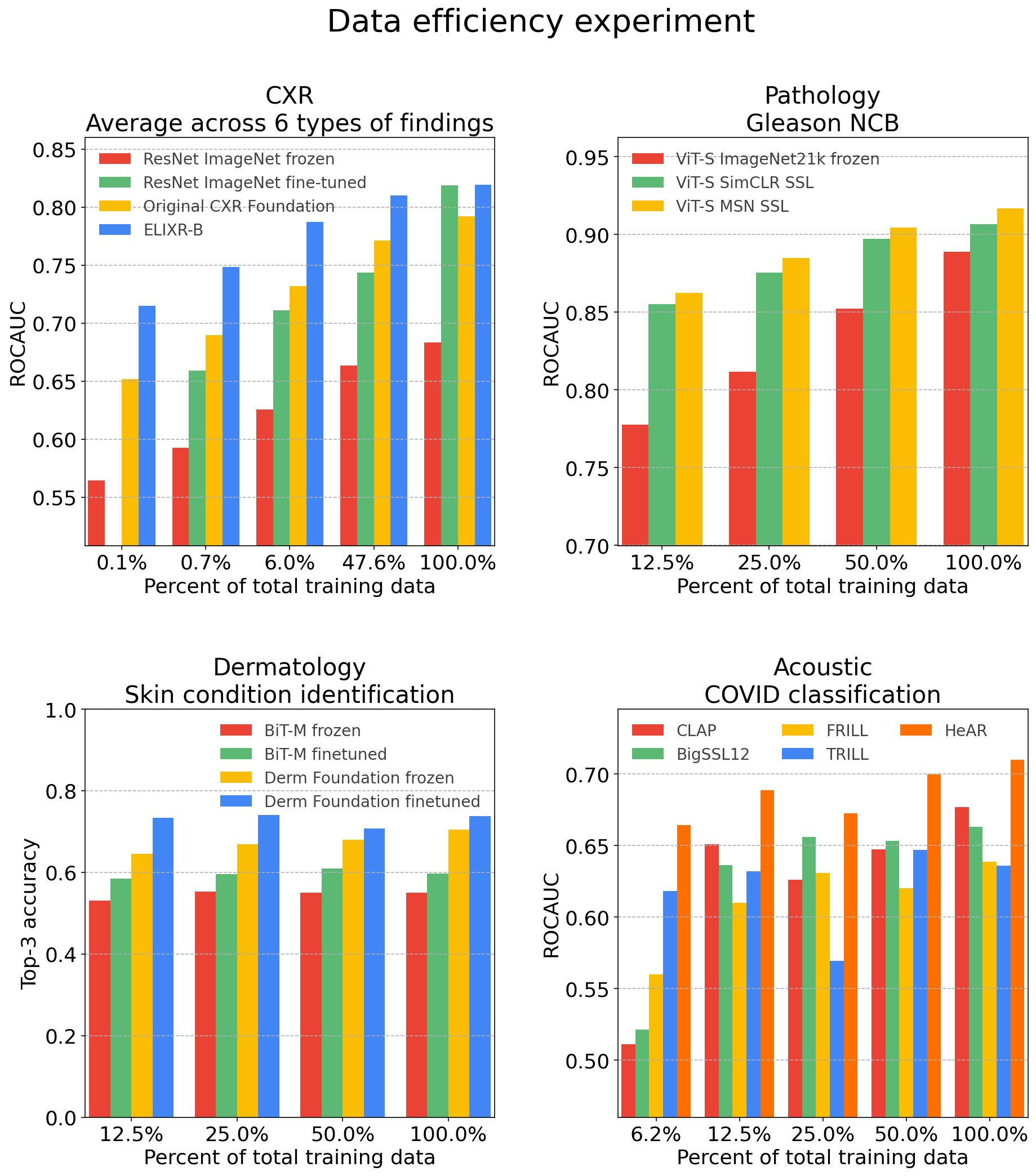}}
  \label{fig:hads_results}
\end{figure}

We evaluated the efficacy of our domain-specific foundation models on a suite of data-efficient classification tasks, benchmarking their performance against generic models when possible. Across various levels of training data subsampling, classifiers leveraging foundation model embeddings consistently outperformed those using generic embeddings, demonstrating superior data efficiency (Figure~\ref{fig:hads_results}).

Beyond data efficiency, our foundation models demonstrate strong generalization capabilities across diverse tasks within their respective domains. The CXR foundation model (ELIXR-B) achieved strong performance on tasks spanning classification, semantic search, visual question answering, and report quality assurance~\citep{Xu2023ELIXR}. The Derm foundation model effectively handles data covering 419 skin conditions, with subgroup analysis revealing no statistically significant performance difference across Fitzpatrick skin types~\citep{Rikhye2024}. HeAR, our health acoustics model, generalizes better when tested on audio recordings from unseen devices, compared to other strong audio encoders (CLAP~\citep{elizalde2023clap}, TRILL~\citep{shor2020towards}, FRILL~\citep{peplinski2020frill}, BigSSL-12~\citep{zhang2022bigssl}), signifying its robust generalization capabilities~\citep{Baur2024}. Finally, the Path foundation model exhibits strong performance on a wide range of tasks encompassing 17 unique tissue types and 12 cancer types, including tumor detection, grading, subtyping, and tissue type classification~\citep{Lai2023}.

These results highlight the potential of our domain-specific foundation models to substantially reduce the data, compute, and technical expertise required for developing task-specific deep learning models in their respective domains, while achieving comparable or superior performance to existing approaches. Further details on each foundational model are presented below.

\subsection{CXR Foundation}

\begin{figure}[htbp]
  \center
  \caption{Performance of ELIXR-C, ELIXR-B, and the original CXR Foundation
embeddings for data-efficient classification. The ROC AUC results of a linear probe are shown averaged across 2 datasets
(CheXpert and Chest X-ray14) for seven findings: atelectasis, cardiomegaly, airspace opacity,
fracture, pneumothorax, consolidation, pleural effusion, and pulmonary edema. Both ELIXR-C
and ELIXR-B demonstrate superior performance compared to the original CXR Foundation at matching dataset
sizes.}
  {\includegraphics[width=4in]{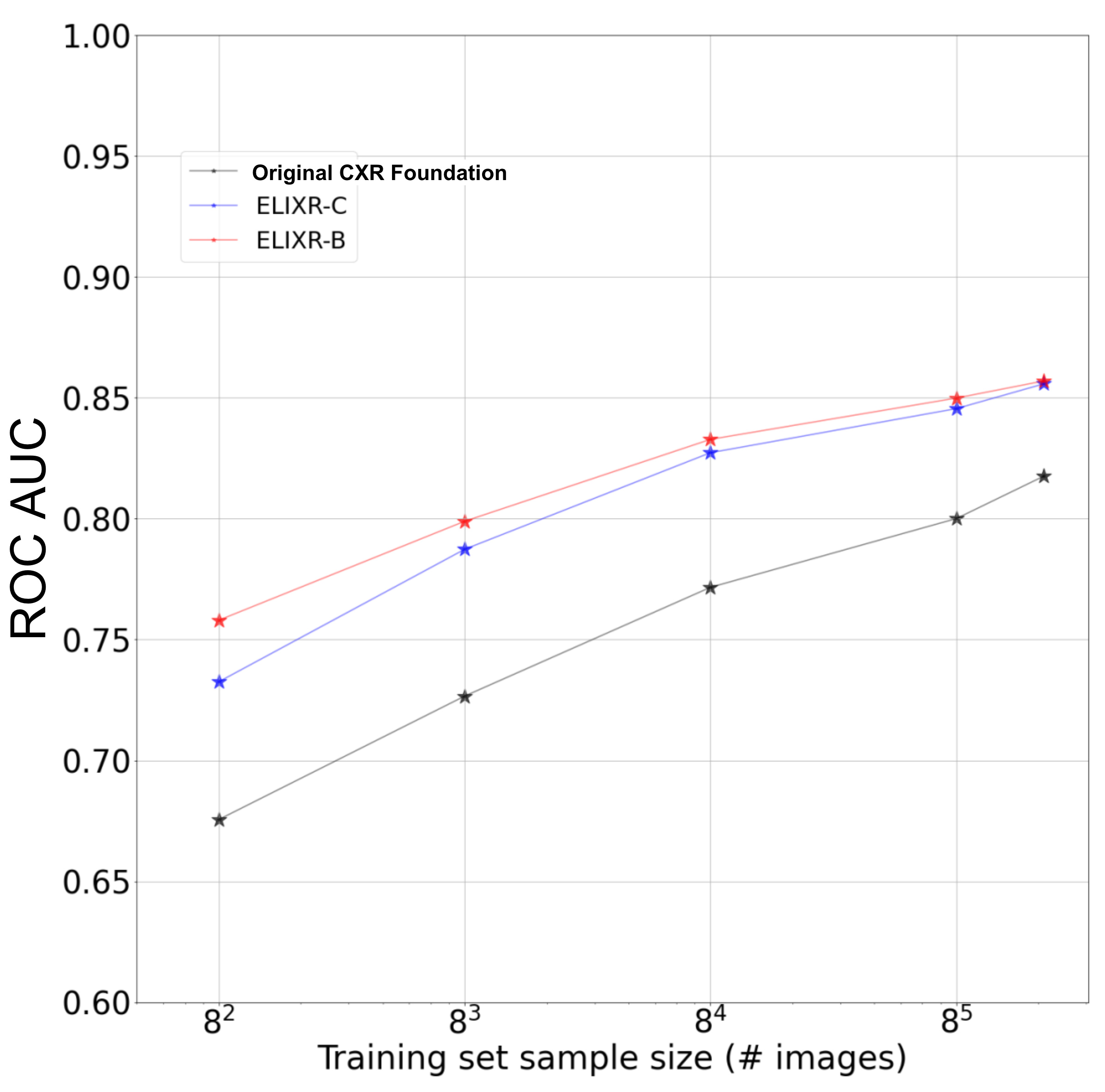}}
  \label{fig:cxrfoundation}
\end{figure}

Area under the curve (ROC AUC) metrics with both linear and non-linear models applied to CXR embeddings were evaluated. On public datasets, such as ChestX-ray14 and CheXpert, results improved the data-accuracy trade-off for models developed across a range of training dataset sizes and several findings. Figure~\ref{fig:cxrfoundation} shows the comparison for all three models using a linear probe. When evaluating the original CXR Foundation's ability to develop tuberculosis models, data efficiency gains were more striking: models trained on the embeddings of just 45 images achieved non-inferiority to radiologists in detecting tuberculosis on an external validation dataset~\citep{Sellergren2022}. For both tuberculosis and severe COVID-19 outcomes, we have shown that non-linear classifiers trained on CXR Foundation embeddings outperformed a model that was fine-tuned on the entire dataset~\citep{Sellergren2022}.

We have also developed a no-code interface to demonstrate the feasibility and performance of binary classification tasks based on CXR Foundation. This demo builds a classifier model using the embeddings from the DICOM images and labels from the ChestX-ray14 dataset, and uses the CXR Foundation to generate image embeddings and train a simple perform binary classification. The demo can also accept DICOMs directly from the browser as input and can be found on \href{https://colab.research.google.com/github/Google-Health/imaging-research/blob/master/cxr-foundation/CXR_Foundation_Interactive_Demo.ipynb}{GitHub.}

\subsection{Path Foundation}

\begin{figure}[htbp]
  \caption{Performance, measured in ROC AUC, of the Path Foundation on 11 histopathology classification tasks via linear probing. The Path Foundation embeddings demonstrate superior performance compared to ImageNet features across all tasks.}
  {\includegraphics[width=\linewidth]{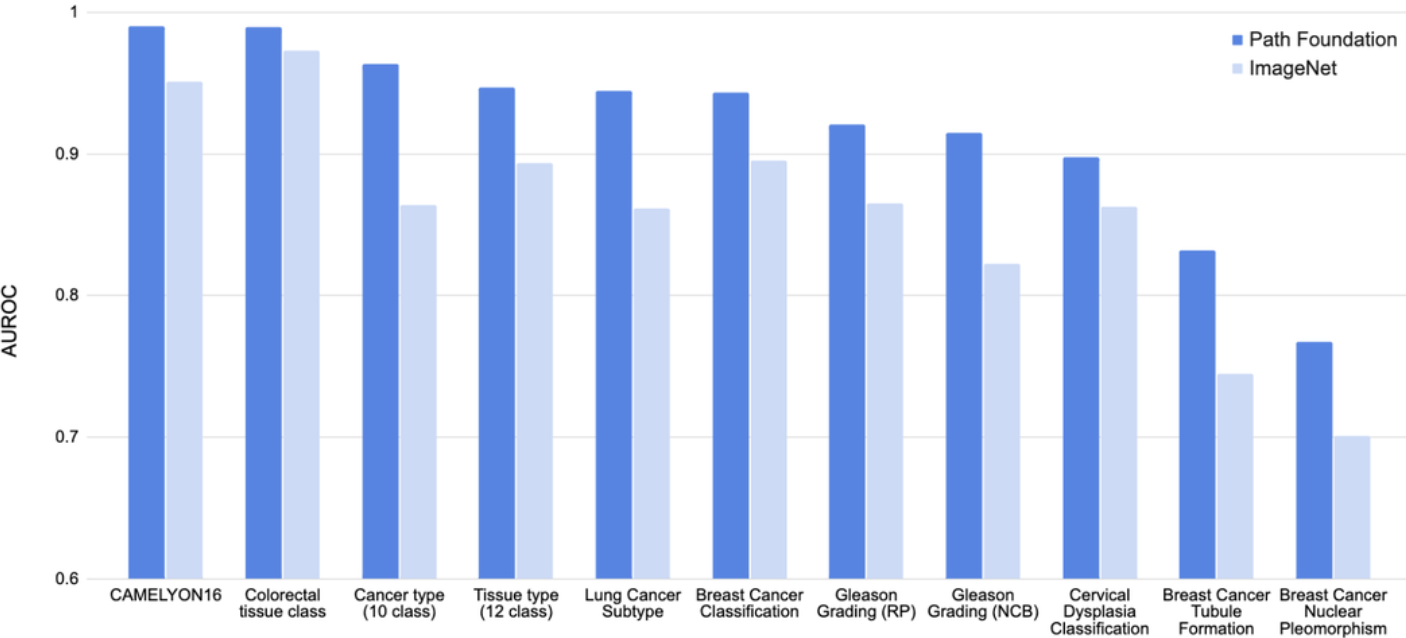}}
  \label{fig:path_results}
\end{figure}

We have evaluated Path Foundation for 11 histopathology tasks via linear probing, shown in Figure~\ref{fig:path_results}. ROC AUC scores are above 0.8 in all but one task.

Besides performance, efficiency is also an important evaluation factor as whole slide digital pathology images are some of the largest medical images (by size in bytes). Recognizing that the efficiency, time and total cost, of embedding generation is a barrier to the development and deployment of pathology ML, we developed a Vertex AI endpoint that generates embeddings for pathology imaging stored both in the Cloud and externally.

All performance measures reported here were performed in triplicate and the mean time required to complete the task is reported. Measurements were done using a Vertex AI endpoint running on n1-highmem-4 virtual machines with an attached NVIDIA Tesla V100 16GB . The endpoint was run as a single node with horizontal scaling disabled.

Embedding generation from the DICOM store and external data were very similar and faster than embedding generation from Google Cloud Storage. The cost per embedding was estimated via \href{https://cloud.google.com/vertex-ai/pricing#n1-series}{Vertex AI pricing} at $\$3.12/\text{hour}$. Table~\ref{table_path2} illustrates estimates for the cost of embedding with the endpoint operated at near capacity. See Section~\ref{app_path} for additional details.


\begin{table}
\centering
\caption{Estimated cost per embedding for the Path Foundation model if Vertex AI embedding endpoint is executed at capacity. Endpoint estimated to cost \$3.12 per hour.}
\label{table_path2}
\begin{tabular}{cccc}
\toprule
 & \textbf{External} & \textbf{Google Cloud} & \textbf{Google Cloud} \\
 & \textbf{Data Source} & \textbf{Storage} & \textbf{DICOM Store} \\
\midrule
\textbf{Embeddings / Hour} & 960,869 & 243,779 & 988,217 \\
\textbf{$\sim$Price (\$) / Embedding} & \$0.000003 & \$0.00001 & \$0.000003 \\
\textbf{$\sim$Embedding / \$} & 307,970 & 78,134 & 316,736 \\
\bottomrule
\end{tabular}
\end{table}

\subsection{Derm Foundation}

\begin{figure}[htbp]
  
  \caption{ROC AUC curves of data efficient classifiers using SCIN dataset for a representative sample of dermatology tasks. (top) shows results of a logistic regression classifier and (bottom) shows results using a simple (two-layer) neural network.
}
  \includegraphics[width=\linewidth]{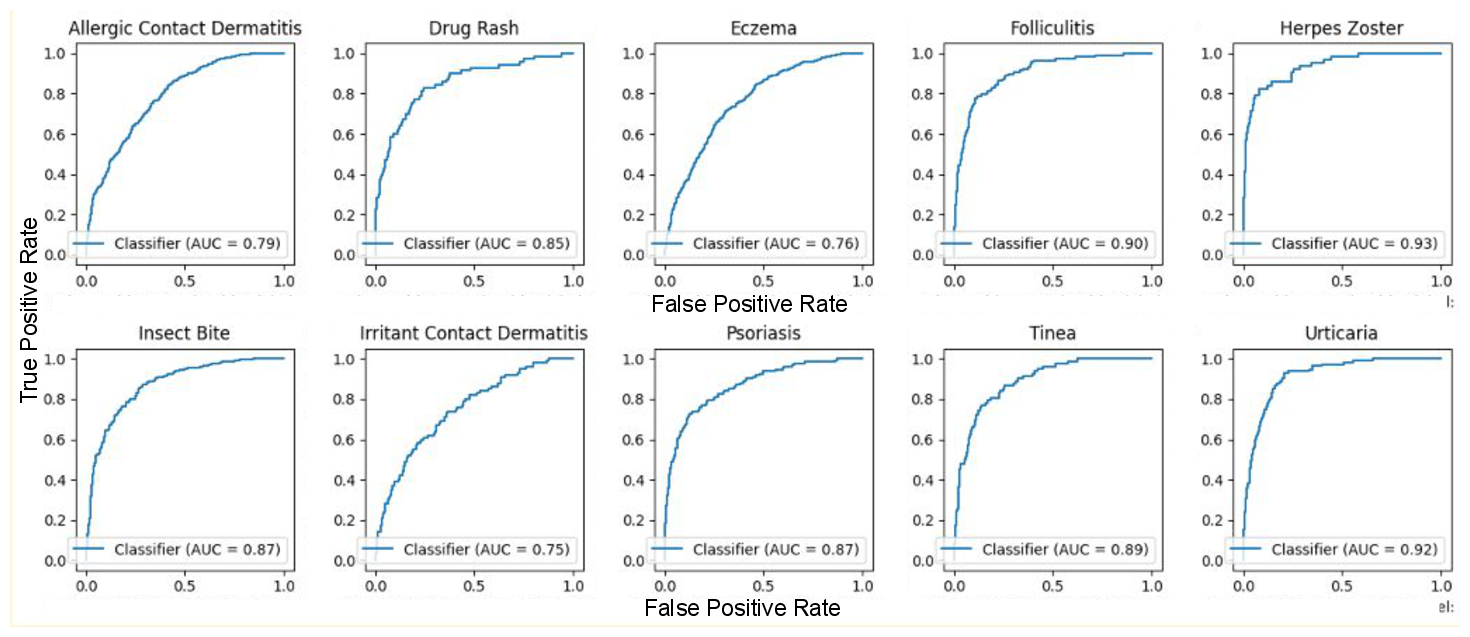}
  \label{fig:derm_results_2}
\end{figure}

To evaluate Derm Foundation, we trained and evaluated classifiers for ten tasks. We used the SCIN (Skin Condition Image Network)~\cite{ward2024crowdsourcing} dataset to train and test these classifiers. SCIN is an open access dataset created by generating representative images from internet users with the goal of providing an accessible diverse dermatology dataset. The results are presented in Figure~\ref{fig:derm_results_2}.

\subsection{HeAR}

HeAR has been trained on a large collection of short (2 seconds, sampled at 16kHz, mono-audio) clips of sounds made by humans, such as coughing, breathing, or speaking. No segmentation or padding was used. The audio typically contains background noise, which helps make the encoder robust to such noise.

By reconstructing masked sounds, it learns meaningful representations that capture underlying patterns useful for downstream tasks. This embedding model can be used to explore the possibility of developing classifiers for health-related non-semantic acoustics. Examples of such tasks include detecting tuberculosis using cough sounds, detecting dementia using non-semantic voice patterns, and detecting air exhalation volume from exhalation sounds.

\begin{figure}[htbp]
  \caption{Comparison of the performance of HeAR to other audio encoders on four cough tasks, based on a single random distribution of train data. The y-axis shows the ROC AUC performance while the x-axis shows the percentage of training data used. HeAR performs favorably across different data regimes and tasks, demonstrating its data efficiency.}
  \includegraphics[width=\linewidth]{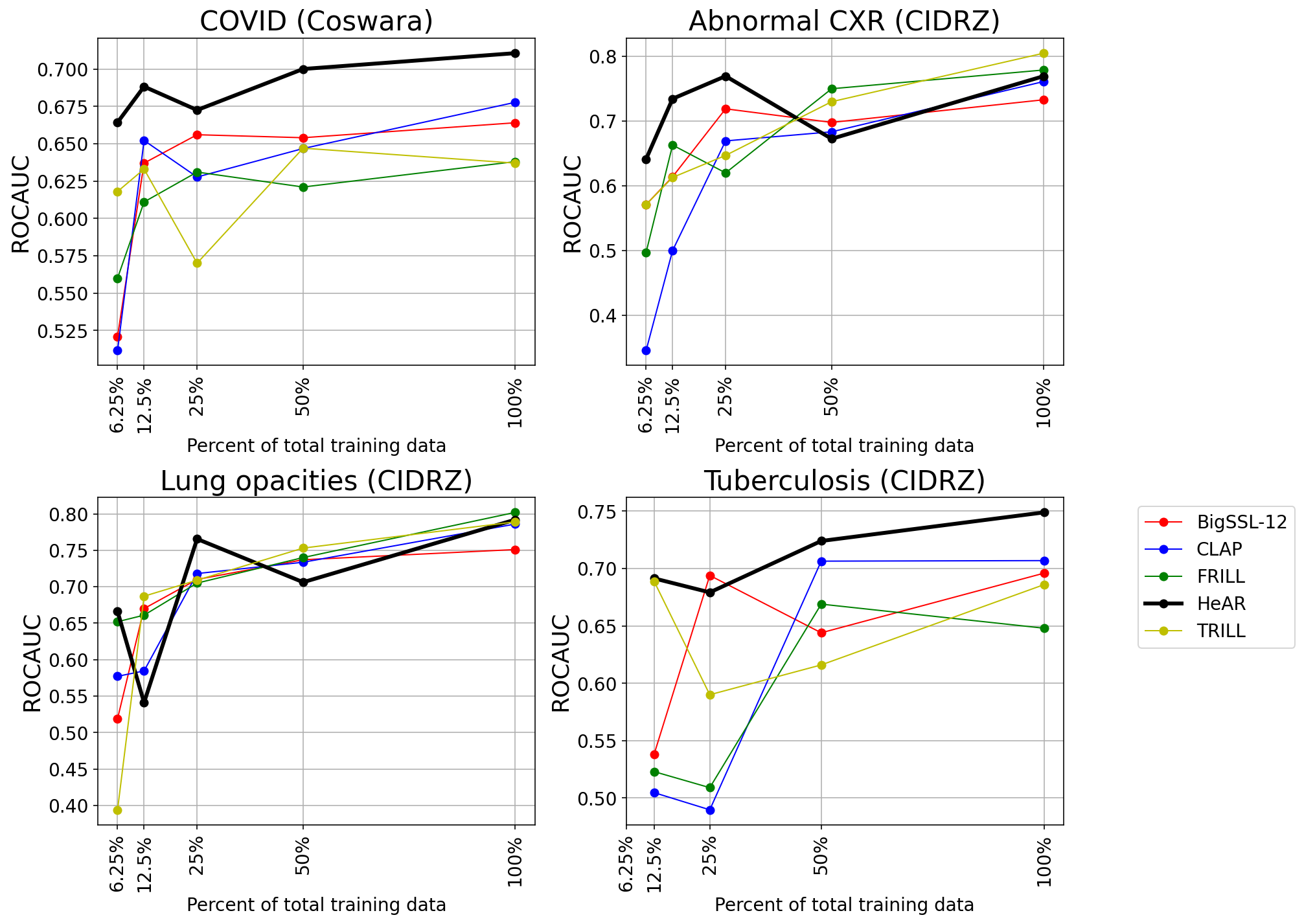}
  \label{fig:hear_results}
\end{figure}

We observed in \cite{Baur2024} that HeAR tends to perform better than available competing audio encoders for such tasks, while typically needing less data, as shown in Figure~\ref{fig:hear_results}. Deployment and test-time scenarios are typically very different from lab tests. For health acoustics, it typically means that the recording devices and data collection protocols can be different. Deep learning models can be brittle to such distribution shifts, and yet we observed that HeAR is more robust to such changes.

\subsection{CT Foundation}

\begin{figure}[htbp]
  \center
  \caption{CT Foundation embeddings performance (ROC-AUC and accuracy) across each task. Each model was trained only using CPUs with different sizes of training data, and the evaluation subset was held constant. N refers to the total size of the evaluation set used to measure performance. Throughout each training run the positive ratio in the training set approximately matches the evaluation set. The lung cancer training and evaluation splits match those of the training and tune sets used in~\citet{ardila2019}.}
  {\includegraphics[width=5in]{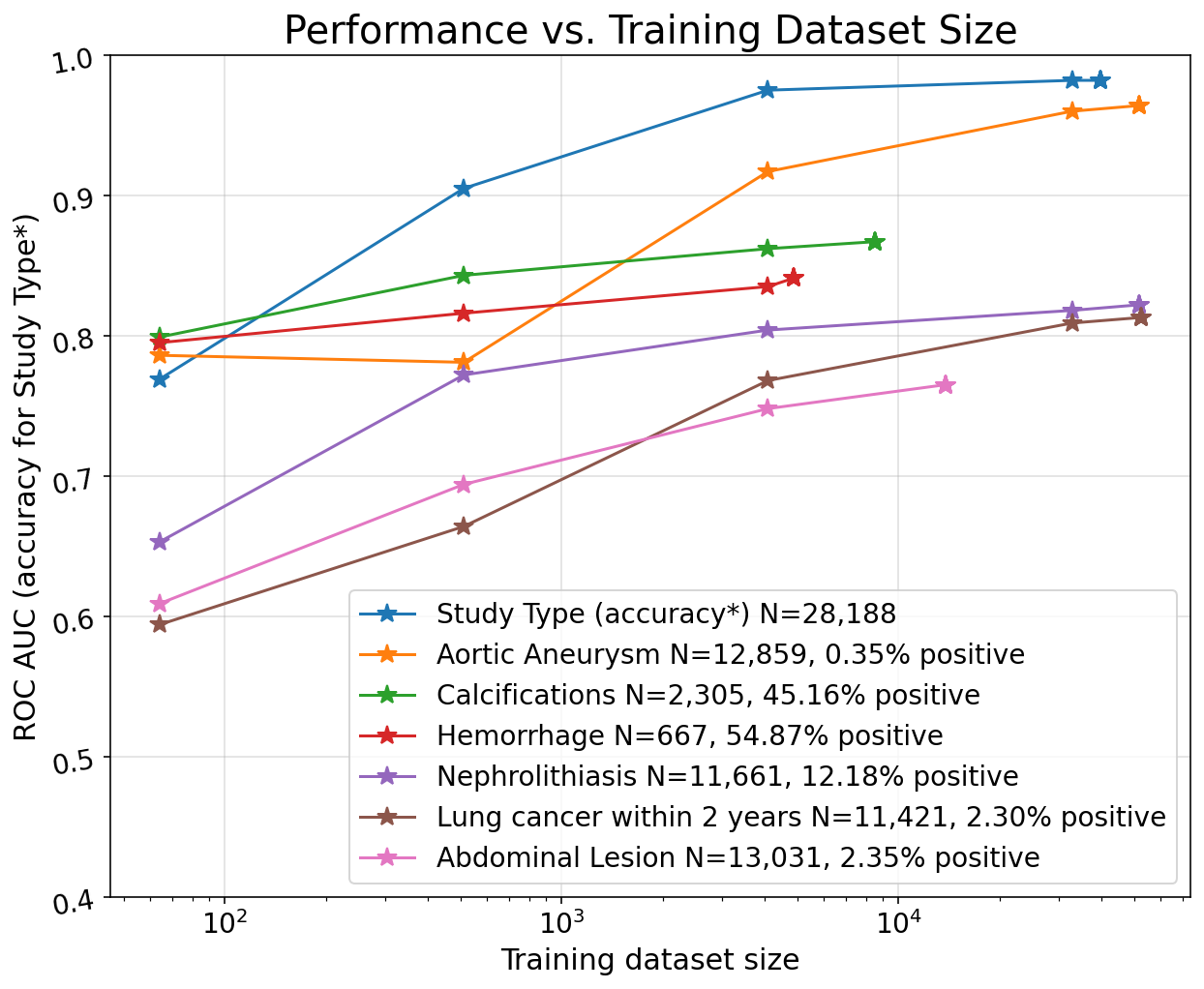}}
  \label{fig:ct_results}
\end{figure}

CT Foundation was trained using three-dimensional computed tomography (CT) scans and corresponding radiology reports across different study types, including Head/Neck, Neck, Spine, Heart, Angiography, Chest, Abdomen and Pelvis, and Extremities. To test CT Foundation’s utility and generalizability, we evaluated the embeddings across seven classification tasks using multilayer perceptron models with increasing amounts of training data. Tasks were diverse, spanning head, chest, and abdominopelvic regions, each involving the detection of abnormalities. These tasks were related to classifying intracranial hemorrhage, calcifications in the chest and heart, lung cancer prediction in the chest, suspicious abdominal lesions, nephrolithiasis, and abdominal aortic aneurysm in abdominopelvic CTs. All binary labels with the exception of the lung cancer and hemorrhage tasks were automatically extracted from the clinical radiology reports, which were written by board certified radiologists. The lung cancer prediction task was drawn from NLST~\citep{nlst2011} and used pathology-confirmed cancer outcomes within 2 years of the lung screening task for cancer positive labels. The hemorrhage task was further verified and labeled by board certified radiologists. The evaluation set to measure the performance was held constant. Figure~\ref{fig:ct_results} shows the results in ROC AUC and accuracy for these tasks. All but one of the ROC AUC measured tasks achieved a score of 0.8 or greater.

\section{Discussion}
\label{sec:discussion}
The ML models within HAI-DEF have been made available for research use over the last 3 years as research endpoints where users can upload images and receive embeddings. HAI-DEF now expands this collection of research endpoints with open-weight models and containerized per-modality solutions that can be deployed as endpoints in user-managed environments for CXR, Path and Derm Foundation. This unlocks use cases beyond research, and enables use on datasets that cannot be processed by a Cloud system due to privacy, institutional policies, or other considerations.

These foundation model research endpoints have been adopted by many researchers, with millions of API calls as of this writing. Users, including both machine learning researchers and clinicians, have utilized these research endpoints to explore a diverse range of applications and have found them valuable for improving performance and building models quickly for research purposes. Use cases advanced by researchers include: using Path Foundation to help distinguish different types of sarcoma at University College London; and using CXR Foundation for identifying necrotizing enterocolitis on neonatal radiology images at Guys St. Thomas Trust.

As we expand the routes by which foundation models can be consumed, it is important to consider various trade-offs. Using model weights directly is the most flexible way to consume the models; for example, the models can be leveraged as part of existing ML software infrastructure or as a constituent model in an ensemble for real time use. However, this requires developers to process the data into a format that models expect (which might differ between different models and modalities). In this regard, consuming the models via endpoints that provide additional preprocessing logic may make it easier to fetch and process data for inference. Deploying endpoints on cloud infrastructure (for instance as a Google Cloud Vertex AI endpoint) also offers scalability without needing researchers or infrastructure staff to manage the complexity of ensuring robustness to fluctuating usage volumes. 

Data use restrictions are another factor that drives differences between the open-weights versus research endpoint approaches. Google-maintained research endpoints can only be used for research scenarios and with de-identified data. This may render it unsuitable for use cases with strict data locality requirements. On the other hand, certain datasets have usage terms that prohibit direct availability of the downstream trained models, or are associated with other sensitivities that might make releasing model weights infeasible. The endpoint approach can respect these restrictions while providing an alternative route to enabling downstream use by other researchers. Finally, another route combines some elements of both open-weights and endpoint approaches: user-deployed endpoints (using the ready-to-deploy containers on Google Cloud’s Model Garden) have the scalability and preprocessing benefits of research endpoints, in addition to meeting the data locality and sensitivity requirements for some users.

With regards to downstream use cases, even though embedding models have been trained on a large diverse datasets (a compute- and data-intensive process), making them useful for specific tasks should always be done via validation and/or tuning on data specific to the problem and patient population of interest. This fine-tuning can be especially important for rarer examples that the embedding models may not have ‘seen’ many of during development. For many applications, we and other researchers observe that using a foundation model can reduce the amount of data needed (i.e., in the low-data regime). In some instances, we additionally find benefits (such as generalization) in a higher data regime. For example, we have found HeAR to display more robust performance when testing on a new audio collection smartphone. 

Reducing bias and promoting fairness and equity are important when using ML in healthcare applications. Fine-tuning with local data specific to the use case, patient population, data acquisition protocol or devices is likely essential to achieving higher performance. Embedding models, by reducing barriers to entry can democratize use of AI in low resource settings. Crucially, \cite{Weng2024AnIA} emphasizes that regardless of the model's origin (a foundation model or otherwise), rigorous evaluation of the final downstream model is necessary to guarantee fairness and mitigate potential biases. We are also looking forward to community feedback and lessons learnt applying these models in a diverse range of tasks. One interesting but challenging goal is to incorporate feedback or additional datasets into the foundation models for improvements and further use by the community. By establishing a positive feedback loop, these models could improve over time and become even more useful for all users. 

Like all ML models, models provided in HAI-DEF have limitations to consider. Though individual details differ, each model in HAI-DEF has been trained on varying amounts of data based on availability; please refer to each associated manuscript for more details. As discussed above, we emphasize the importance of rigorous evaluation per use case and population, to improve overall performance and address issues of bias, fairness, and equity. Finally, while we strove to make the models useful for a variety of use cases, some applications may need further development. For example, the models were developed with a focus on classification tasks, and prognosis tasks will need to be further evaluated. Image segmentation and generation tasks are also currently not supported. Further, specific requirements such as smaller models (e.g. for on-device applications on a mobile device) or lower latency will need other techniques such as distillation to the target model size of interest.

\section{Future Work}
\label{sec:future}
We hope to expand and improve the HAI-DEF program to make it useful for an even wider range of applications and to accelerate the adoption of AI in healthcare. One such example is the adoption of open language models in applications working with health guidelines. We plan to release a health guidelines toolkit containing code recipes to help developers evaluate model performance and incorporate them into applications such as training and education.

We look forward to receiving feedback and success stories from the community, both in terms of how existing models can be improved but also whether they can be expanded to support use cases we may not have considered yet.

\section{Acknowledgements}
\label{sec:acknowledgements}
HAI-DEF is the result of extensive collaborative efforts involving numerous individuals and institutions. We extend our sincere gratitude to the NIH Clinical Center, Stanford Medical Center, Project Coswara, CoughVID, and the Center for Infectious Disease Research in Zambia for their collaboration and for making their datasets available to the research community. We also thank DeepHealth and Apollo Radiology for their expertise and datasets. Additionally, the models and results are in part based upon data generated by the TCGA Research Network and the National Lung Screening Trial (NLST)~\cite{nlst2011}.

We gratefully acknowledge the contributions of the Google Research and DeepMind teams, particularly their work on software infrastructure and modeling. We further acknowledge Google Cloud for making these research endpoints available to the research community, enabling broader access and facilitating further advancements. We are indebted to all the clinicians who have contributed their time and expertise in labeling data and evaluating these models. Their contributions have been crucial to the development and validation of our work. Finally, the authors thank Jonathan Krause and Dale Webster for their review and feedback.

\clearpage
\bibliography{references}

\newpage
\appendix
\section{Appendix}
\label{sec:appendix}
\subsection{Background and Methodology For Path Foundation Evaluations}
\label{app_path}
\begin{figure}[htbp]
  \centering
  \caption{Illustration of patch embeddings computed across a histopathology image. Areas of the image with same coloring exhibited similar $k$-means clustering from patch level embeddings.}
  {\includegraphics[width=3in]{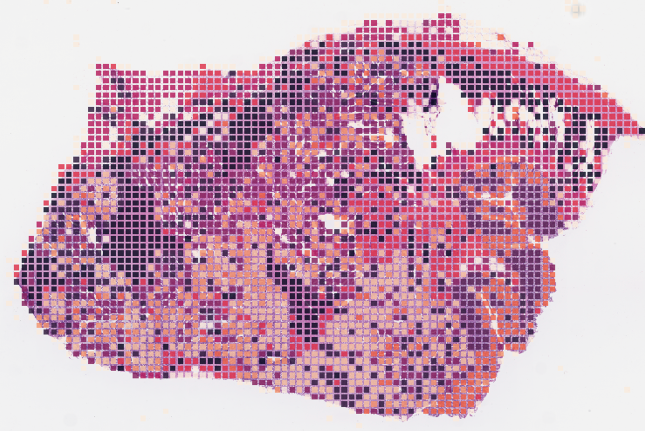}}
  \label{fig:path_patches}
\end{figure}

At its highest magnification, digital pathology images are gigapixel-sized, measuring in the tens to hundreds of thousands of pixels per slide. The Path embeddings model has been trained to produce high quality machine learning embeddings for image patches of size $224~\times~224$ pixels cropped from pathology images. At high magnification, these image patches represent a relatively small portion of the slide, and enable the representation of intra-slide heterogeneity in different parts of the tissue specimen. However, computing patch embeddings across such high magnification pathology images can require the generation of tens of thousands of embeddings for a single whole slide image (Figure~\ref{fig:path_patches}).

The endpoint that was used for evaluations runs as a deployable online prediction REST service within Vertex AI and can generate embeddings for data stored both within Cloud (Cloud DICOM store or Google Cloud Storage) and from external data sources (i.e., sending the data as part of the request). The REST service supports requests that may contain multiple patch embeddings per-data-source and/or multiple data sources. As with all Vertex AI prediction services the service can be configured to horizontally scale to meet demand. To generate embeddings, the endpoint will: 1) retrieve the imaging necessary to complete the embedding generation, 2) generate the embedding, and 3) return the embeddings to the caller. For Cloud data sources, it is optimal to co-locate the Vertex AI endpoint hosting embedding generation in the same data center that hosts the images in order to improve data transport efficiency.

When embeddings are generated from data not stored in Cloud, data are sent directly within the request as a base64 encoded compressed image. This increases the amount of imaging data that can fit within the request. The endpoint supports decoding commonly used compression methods (e.g., lossy JPEG, PNG for lossless data transmission, etc.). Regardless of exactly how imaging is encoded, on receipt, the endpoint extracts the pixel data directly from the request and then generates and returns the embedding result. Multiple factors affect the time required to generate embeddings and the associated cost, including: the compute (CPU and GPU) that backs the Vertex AI embedding generation service, the data source (e.g., DICOM store, Google Cloud Storage, or external data), the image compression format, and the total size of the embedding request.

DICOM embeddings were generated from non-overlapping regions of a DICOM VL Whole Slide Microscopy Image. The DICOM instance was encoded using: $256 \times 256$ pixel frames that were encoded using JPEG baseline transfer syntax and \texttt{TILED\_FULL} organization. Google Cloud Storage embeddings were generated from JPEG images saved to Google Cloud Storage. Image dimensions matched the embedding model patch dimensions ($224~\times~224$ pixels) to enable stored images to be used directly as input to the embedding model. For the external data, the images stored in Google Cloud Storage were read in their entirety and used as the source images for the tests. Differences in utilization stemmed mainly from data retrieval optimizations implemented for Google Cloud DICOM stores using EZ-WSI DICOMWeb~\cite{ez_wsi2023}.

\begin{table}
\centering
\caption{Vertex AI endpoint performance metrics subjected to 10 concurrent requests for 5,000 image embeddings.}
\label{table_path1}
\begin{tabular}{cccc}  
\toprule 
  & \textbf{External} & \textbf{Google Cloud} & \textbf{Google Cloud} \\ 
  & \textbf{Data Source} & \textbf{Storage} & \textbf{DICOM Store} \\ 
\midrule 
\textbf{CPU Utilization (\%)} & 123 & 122 & 93 \\ 
\textbf{GPU Duty Factory (\%)} & 27 & 7 & 25 \\ 
\bottomrule 
\end{tabular}
\end{table}

To estimate the total maximum capacity of an embedding endpoint for the various imaging data sources it was empirically determined that performance metrics, CPU utilization, and GPU duty factor plateaued when the embedding endpoint was subjected to 10 parallel requests for 5,000 embeddings; metrics shown in Table~\ref{table_path1}. The total time required to generate 50,000 embeddings by 10 parallel processes was quantified. Total time was defined as the time between sending the first request from the client to reception of embeddings.

\end{document}